\documentclass[11pt,a4paper]{article}
\usepackage[hyperref]{eacl2021}
\usepackage{times}
\usepackage{latexsym}

\usepackage{booktabs}
\usepackage{graphicx}
\usepackage{amsmath}
\usepackage{rotating}
\newcommand\ang{70}

\usepackage{microtype}

\aclfinalcopy 

\usepackage{lipsum}

\newcommand\blfootnote[1]{%
  \begingroup
  \renewcommand\thefootnote{}\footnote{#1}%
  \addtocounter{footnote}{-1}%
  \endgroup
}

\def\subtxt#1{\ensuremath{{}_{\textnormal{#1}}}}

\title{When Do You Need Billions of Words of Pretraining Data?}

\author{Yian Zhang,$^{*,1}$ Alex Warstadt,$^{*,2}$ Haau-Sing Li,$^{3}$ and Samuel R. Bowman$^{1,2,3}$ \\
 $^{1}$Dept. of Computer Science, $^{2}$Dept. of Linguistics, $^{3}$Center for Data Science\\
New York University\\
\{yian.zhang, warstadt, xl3119, bowman\}@nyu.edu
}

\date{}

\begin{document}
\maketitle
\begin{abstract}
NLP is currently dominated by general-purpose pretrained language models like RoBERTa, which achieve strong performance on NLU tasks through pretraining on billions of words. But what exact knowledge or skills do Transformer LMs learn from large-scale pretraining that they cannot learn from less data? We adopt four probing methods---classifier probing, information-theoretic probing, unsupervised relative acceptability judgment, and fine-tuning on NLU tasks---and draw learning curves that track the growth of these different measures of linguistic ability with respect to pretraining data volume using the MiniBERTas, a group of RoBERTa models pretrained on 1M, 10M, 100M and 1B words. We find that LMs require only about 10M or 100M words to learn representations that reliably encode most syntactic and semantic features we test. A much larger quantity of data is needed in order to acquire enough commonsense knowledge and other skills required to master typical downstream NLU tasks. The results suggest that, while the ability to encode linguistic features is almost certainly necessary for language understanding, 
it is likely that other forms of knowledge are the major drivers of recent improvements in language understanding among large pretrained models.

\blfootnote{*Equal Contribution}

\end{abstract}

\section{Introduction}

\begin{figure}
    \centering
    \includegraphics[width=\columnwidth]{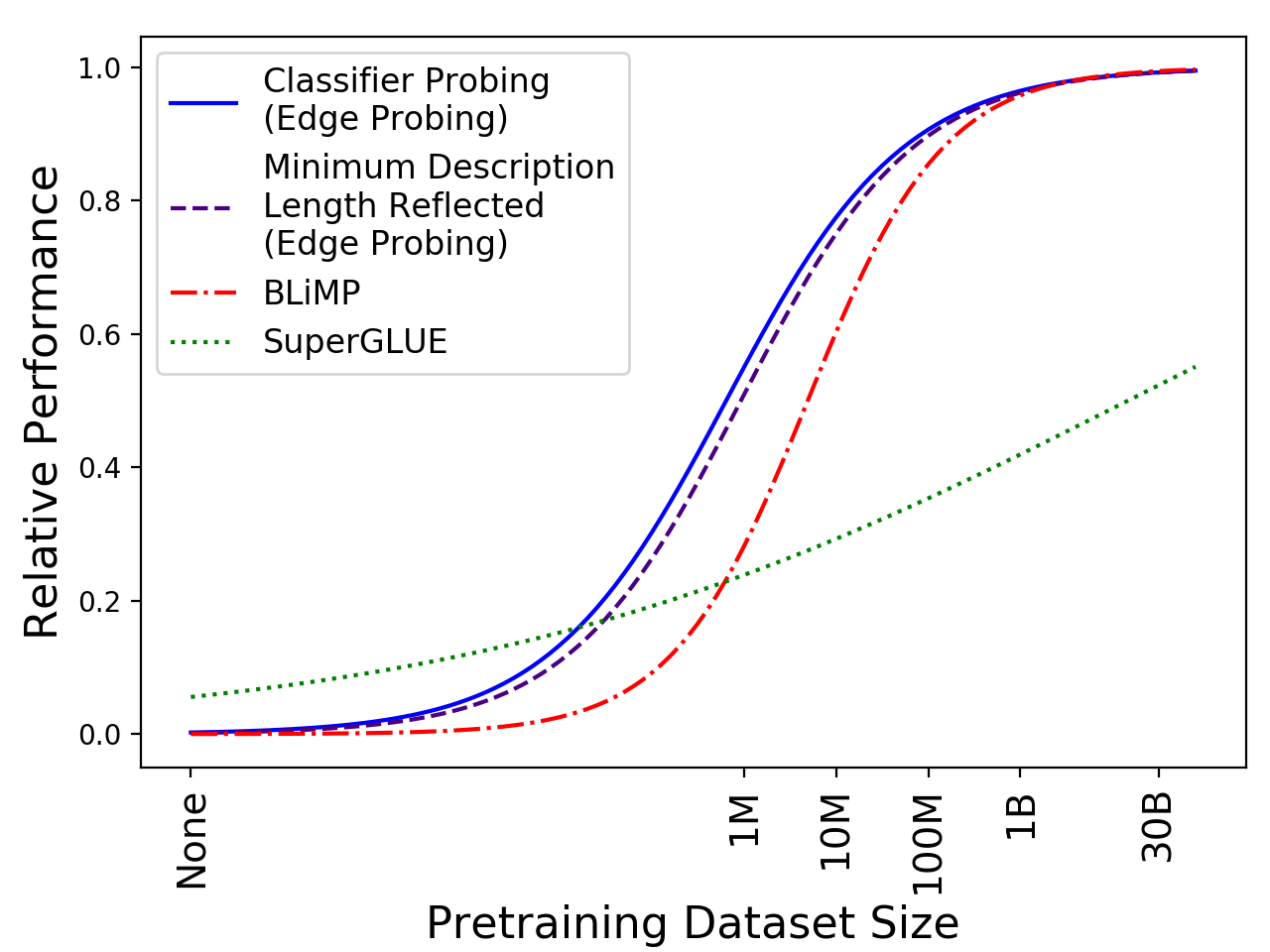}
    \caption{Overall learning curves for the four probing methods. For each method, we compute overall performance for each RoBERTa model tested as the macro average over sub-task's performance after normalization. We fit a logistic curve which we scale to have a maximum value of 1.}
    \label{fig:overall}
\end{figure}

Pretrained language models (LMs) like BERT and RoBERTa have become ubiquitous in NLP. These models use massive datasets on the order of tens or even hundreds of billions of words \cite{brown2020gpt3} to learn linguistic features and world knowledge, and they can be fine-tuned to achieve good performance on many downstream tasks. 

Much recent work has used probing methods to evaluate what these models have and have not learned \citep{Belinkov2019AnalysisMI,tenney2019you,rogers2020primer,ettinger2020bert}. Since most of these works only focus on models pretrained on a fixed data volume (usually billions of words), many interesting questions regarding the effect of the amount of pretraining data remain unanswered: What do data-rich models know that models with less pretraining data do not? How much pretraining data is required for LMs to learn different grammatical features and linguistic phenomena? Which of these skills do we expect to improve if we increase the pretraining data to over 30 billion words? Which aspects of grammar can be learned from data volumes on par with the input to human learners, around 10M to 100M words \cite{hart1992parenting}?

With these questions in mind, we probe the MiniBERTas \citep{warstadt-etal-2020-learning}, a group of RoBERTa models pretrained on 1M, 10M, 100M, and 1B words, and RoBERTa\subtxt{BASE}\citep{liu2019roberta} pretrained on about 30B words, using four methods: 
First we use standard \emph{classifier probing} on the edge probing suite of NLP tasks \citep{tenney2019you} to measure the quality of the syntactic and semantic features that can be extracted by a downstream classifier with each level of pretraining. Second, we apply \emph{minimum description length probing} \citep{voita2020informationtheoretic} to the edge probing suite, with the goal of quantifying the accessibility of these features. Third, we probe the models' knowledge of various syntactic phenomena using unsupervised acceptability judgments on the BLiMP suite \citep{warstadt2020blimp}. Fourth, we fine-tune the models on five tasks from SuperGLUE \citep{wang2019superglue}, to measure their ability to solve conventional NLU tasks. 

Figure \ref{fig:overall} shows the interpolated learning curves for these four methods as a function of the amount of pretraining data. We have two main findings: First, the results of three probing methods we adopt show that the linguistic knowledge of RoBERTa pretrained on 100M words is already very close to that of RoBERTa\subtxt{BASE}, which is pretrained on around 30B words. Second, RoBERTa requires billions of words of pretraining data to make substantial improvements in performance on dowstream NLU tasks. From these results, we conclude that there are skills critical to solving downstream NLU tasks that LMs can only acquire with billions of words of pretraining data and that we need to look beyond probing for linguistic features to explain why LMs improve at these large data scales.

\section{Methods}
\begin{figure*}[t]
    \centering
    \includegraphics[width=\textwidth]{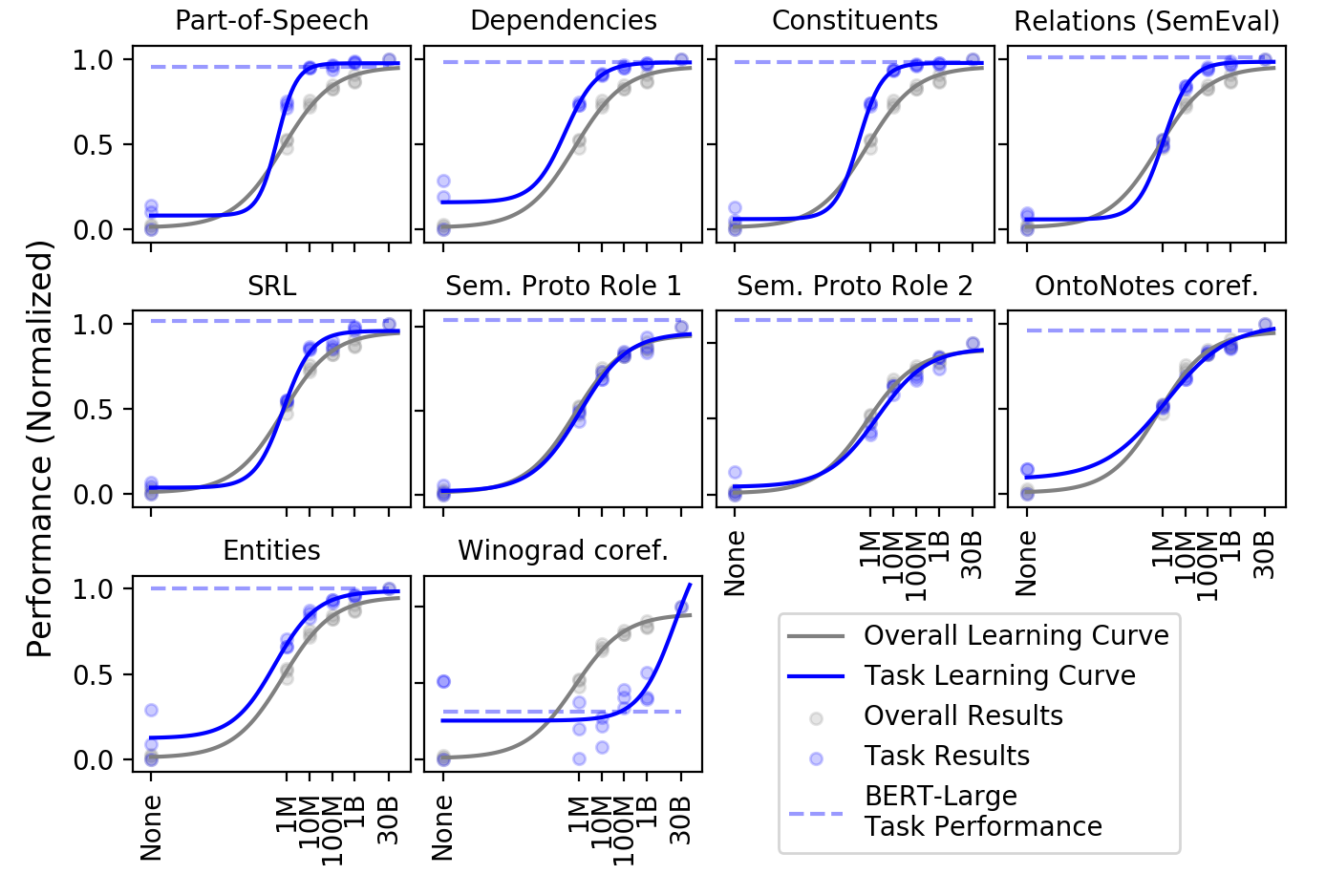}
    \caption{Classifier probing results for each task in the edge probing suite, adjusted using min-max normalization. The overall results are identical in each subplot, and are repeated to make comparisons easier. For context, we also plot BERT\subtxt{LARGE} performance for each task as reported by \citet{tenney-etal-2019-bert}.}
    \label{fig:edge_probing}
\end{figure*}

\begin{figure}[t]
    \centering
    \includegraphics[width=\columnwidth]{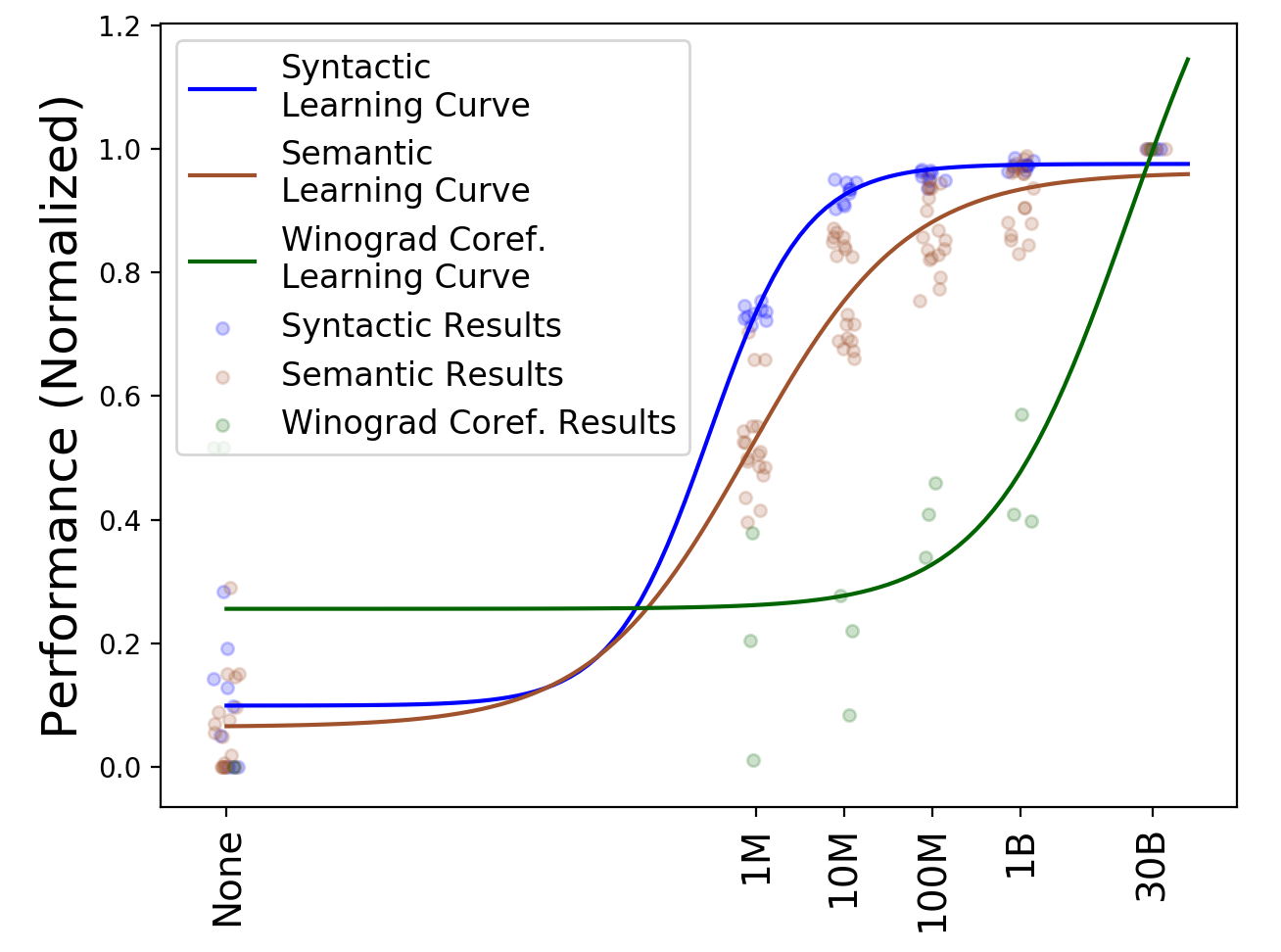}
    \caption{Edge Probing results for each group of tasks adjusted using min-max normalization. Syntactic tasks are Part-of-Speech, Dependencies, and Constituents. The commonsense task is Winograd coref. Semantic tasks are all remaining tasks.}
    \label{fig:edge_probing_syntactic_vs_semantic}
\end{figure}

We probe the MiniBERTas, \footnote{\href{https://huggingface.co/nyu-mll}{\url{https://huggingface.co/nyu-mll}}.} a set of 12 RoBERTa models pretrained from scratch by \citet{warstadt-etal-2020-learning} on 1M, 10M, 100M, and 1B words sampled from a combination of Wikipedia and Smashwords, the sources that \citet{devlin2019bert} use to pretrain BERT, and a subset of those used for RoBERTa. \citeauthor{warstadt-etal-2020-learning} ran pretraining 25 times with varying hyperparameter values for each of 1M, 10M, and 100M, and 10 times for 1B. For each dataset size, they released the three models with the lowest dev set perplexity, yielding 12 models in total.

We also test the publicly available RoBERTa\subtxt{BASE}
\footnote{ \href{https://github.com/pytorch/fairseq/tree/master/examples/roberta}{\url{https://github.com/pytorch/fairseq/tree/master/examples/roberta}}} 
\cite{liu2019roberta}, which is pretrained on about 30B words,\footnote{In addition to Wikipedia and Smashwords, RoBERTa\subtxt{BASE} is also trained on news and web data.} and 3 RoBERTa\subtxt{BASE} models with randomly initialized parameters.

We probe the MiniBERTas using four methods: classifier probing on the edge probing suite, minimum description length probing on the edge probing suite, unsupervised acceptability judgments on BLiMP, and fine-tuning on NLU tasks from SuperGLUE.\footnote{The code for all four experiments can be found at \href{https://github.com/nyu-mll/pretraining-learning-curves}{\url{https://github.com/nyu-mll/pretraining-learning-curves}}.} In each probing experiment, we test all 16 models on each task involved. For all experiments except for BLiMP, we use min-max normalization to adjust the results into the range of [0, 1], where 0 represents the worst score of any model on the task (usually a randomly initialized one), and 1 represents the best score of any model (usually RoBERTa$\subtxt{BASE}$).\footnote{The unnormalized results are included in the appendix.} We plot the results in a figure for each task, where the $y$-axis is the (normalized) score and the $x$-axis is the amount of pretraining data.\footnote{We plot the no-pretraining random baseline with an $x$-value of 1.} To show the overall trend of improvement, we use non-linear least squares to fit a logistic function to the points after log transforming the $x$-values.\footnote{We assume log-logistic learning curves because of the goodness of the fit to our empirical findings. It may also be reasonable to fit an exponential learning curve \cite{heathcote2000power}.}

\section{Classifier Probing}

We use the widely-adopted probing approach of \citet{ettinger2016probing}, \citet{adi2017fine}, and others---which we call  \emph{classifier probing}---to test the extent to which linguistic features like part-of-speech and coreference are encoded in the MiniBERTa representations. In these experiments we freeze the representations and train MLP classifiers for the ten probing tasks in the edge probing suite \citep{tenney2019you}.\footnote{Task data source: Part-of-Speech, Constituents, Entities, SRL, and OntoNotes coref. from \citet{weischedel2013ontonotes}, Dependencies from \citet{silveira-etal-2014-gold}, Sem. Proto Role 1 from \citet{AAAI1714997}, Sem. Proto Role 2 from \citet{rudinger-etal-2018-neural-davidsonian}, Relations (SemEval) from \citet{hendrickx-etal-2010-semeval}, Winograd coref. from \citet{rahman-ng-2012-resolving,white2017}} 

Admittedly, classifier probing has recently come under scrutiny. \citet{hewitt2019designing} and \citet{voita2020informationtheoretic} caution that the performance achieved in the classifier probing setting reflects a combined effort of the representations and the probe, so a probing classifier's performance does not precisely reveal the quality of the representations. However, we think it is still valuable to include this experiment setting for two reasons: First, the downstream classifier setting and F1 evaluation metric make these experiments easier to interpret in the context of earlier results than results from relatively novel probing metrics like minimum description length. Second, we focus on relative differences between models rather than absolute performance and include a randomly initialized baseline model in the comparison. When the model representations are random, the probe's performance reflects the probe's own ability to solve the target task. Therefore, any improvements over this baseline value are due to the representation rather than the probe itself. On the other hand, since other probing methods are well motivated, we 
we also look to minimum description length probing \cite{voita2020informationtheoretic} in the next section to quantify not just how well a probe can perform, but how complex the probe is.


\paragraph{Task formulation and training}

Following \citeauthor{tenney2019you}, we take the input for each task to be a pair of token spans or a single span of tokens. For each task $T$, if $T$ is a pairwise task, we train two attention pooling functions $f_T^1$ and $f_T^2$, and for each span pairs $(S_i^1,S_i^2)$ we generate a representation pair $(r_i^1,r_i^2)=(f_T^1(S_i^1),f_T^2(S_i^2))$. Then for each label $L_j$ of $T$, the probe (which is an MLP) takes in $(r_i^1,r_i^2)$ and performs a binary classification to predict whether $L_j$ is the correct label. For tasks that involve only a single span (Part-of-Speech, Constituents, and Entities), $S_i^2$ and $f_T^2$ are omitted. We adopt the `mix' representation approach, so each token representation $(t_i^k)_p$ from $S_i^k=\{(t_i^k)_0, (t_i^k)_1, ...\}$ is a linear combination of RoBERTa's layer activations projected to a 256-dimensional space.


 For each task, we fix validation interval to be 1000 steps, early stopping patience to be 20 steps, learning rate patience to be 5 steps, and sample 5 combinations of batch size and learning rate randomly\footnote{The search range for batch size and learning rate are \{8,16,32,64\} and \{5e-5,1e-4,5e-4\} respectively.} to tune the model with the lowest MLM perplexity at each pretraining scale using the Adam optimizer \citep{kingma2014adam}. We use the best hyperparameter setting to train all the models of that scale on the task.

\paragraph{Results}

We plot the experiment results in Figure \ref{fig:edge_probing}, and in each subplot we also plot the overall edge-probing performance, which we calculate for each MiniBERTa as its average F1 score on the 10 edge-probing tasks (after normalization).


From the single-task curves we conclude that most of the feature learning occurs with $<$100M words of pretraining data. Based on the best-fit logistic curve, we can estimate that 90\% of the attainable improvements in overall performance are achieved with $<$20M words. Most plots show broadly similar learning curves, which rise sharply with less than 1M words of pretraining data, reach the point of fastest growth around 1M words, and are nearly saturated with 100M words. The most notable exception to this pattern is the Winograd task, which only rises significantly between 1B and 30B words of pretraining data.\footnote{These results are also somewhat more noisy due to well-known idiosyncrasies of this task.} As the Winograd task is designed to test commonsense knowledge and reasoning, we infer that these features require more data to encode than syntactic and semantic ones.

There are some general differences that we can observe between different types of tasks. Figure \ref{fig:edge_probing_syntactic_vs_semantic} shows the aggregated learning curves of syntactic, semantic, and commonsense tasks. The syntactic learning curve rises slightly earlier than the semantic one and 90\% of the improvements in syntactic learning can be made with about 10M words, while the semantic curve is still rising slightly after 100M. This is not surprising, as semantic computation is generally thought to depend on syntactic representations \citep{heim1998semantics}, and \citet{tenney-etal-2019-bert} report a similar result. The commonsense learning curve (for Winograd coref.~only) clearly rises far later, and is projected to continue to rise long after syntactic and semantic features stop improving.

\section{Minimum Description Length Probing}
\begin{figure*}[t!]
    \centering
    \includegraphics[width=\textwidth]{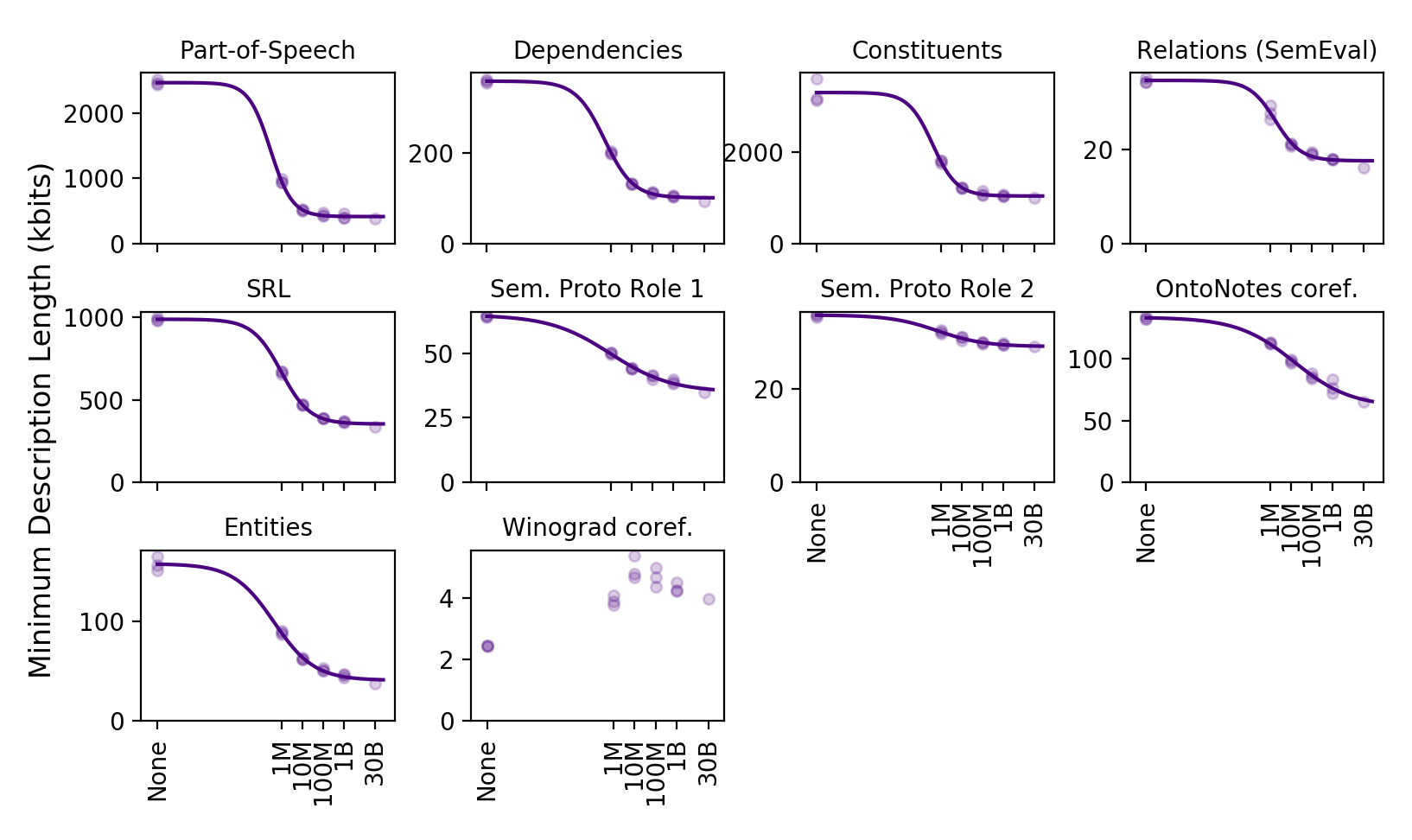}
    \includegraphics[width=\textwidth]{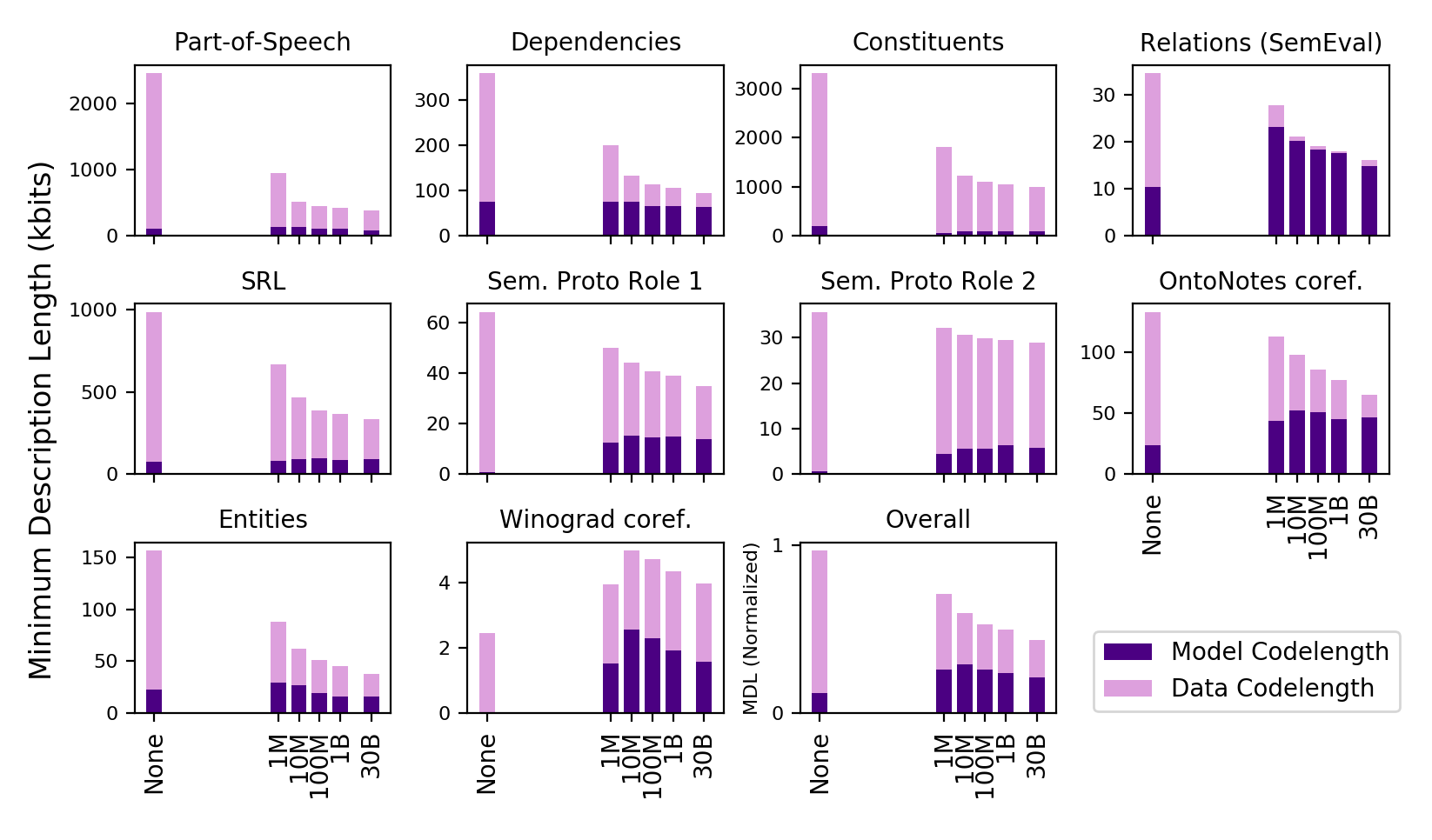}
    \caption{MDL results for each edge probing task. We do not plot a logistic curve for the Winograd coref.~results because would could not find an adequate fit.}
    \label{fig:mdl}
\end{figure*}



In this experiment, we study the MiniBERTas with minimum description length (MDL) probing  \citep{voita2020informationtheoretic}, with the goal of revealing not only the total amount of feature information extracted by the probe, but also the efforts taken by the probe to extract the features. MDL measures the minimum number of bits needed to transmit the labels for a given task given that both the sender and the receiver have access to the pretrained model's encoding of the data. In general, it is more efficient for the sender not to directly transmit the labels, but to instead transmit a decoder model that can be used to extract the labels from the representations. If a decoder cannot losslessly recover the labels, then some additional information must be transmitted as well. In this way, fewer bits are required to transmit the same information, i.e.~the data is \emph{compressed}. 

The MDL of a dataset for an encoder model is thus a sum of the estimates of two terms: The data codelength is the number of bits needed to transmit the labels assuming the receiver has the trained decoder model, i.e.~the cross-entropy loss of the decoder. The model codelength is the number of bits needed to transmit the decoder parameters. There is a tradeoff between data codelength and model codelength: A simpler decoder is likely to have worse performance (i.e. decreasing model codelength often increases data codelength), and vice-versa.



We adopt \citeauthor{voita2020informationtheoretic}'s \emph{online code} estimation of MDL. We compute the online code by partitioning the training data into 11 portions: $\{(x_j, y_j)\}_{j=t_{i-1}+1}^{t_{i}}$ for $1{\leq}i{\leq}11$. The values of $t_0, ..., t_{11}$ are the numbers of examples corresponding to the following proportions of the training data: 0\%, 0.1\%, 0.2\%, 0.4\%, 0.8\%, 1.6\%, 3.2\%, 6.25\%, 12.5\%, 25\%, 50\%, 100\%. Then for each $i{\in}[1,10]$, we train an MLP with parameters $\theta_i$ on portions $1$ through $i$, and compute its loss on portion $i{+}1$. Finally we compute the online codelength as the sum of the ten loss values and the codelength of the first data portion under a uniform prior:

\begin{align}
\nonumber
& L^{online}(y_{1:n}|x_{1:n}) = t_1\log_2K - \\
& -\sum\limits_{i=1}^{10}\log_2p_{\theta_i}(y_{t_{i}+1:t_{i+1}}|x_{t_{i}+1:t_{i+1}}).
\label{eq:online_code}
\end{align} where the number of labels $K=2$ for all edge probing tasks.

In \citeauthor{voita2020informationtheoretic}'s MDL experiments with the edge probing suite, the authors convert the edge probing tasks to multi-class classification problems. Our implementation skips this step and follows the \citeauthor{tenney2019you} edge probing task formation with one classifier head per candidate label, so that we can include the tasks that involve multiple correct labels, enabling a full comparison between our MDL results and our conventional edge probing results. Since \citeauthor{voita2020informationtheoretic} report that MDL is stable across reasonable hyperparmeter settings, we use the settings described in \citet{tenney2019you} for all the models we probe.
 
\paragraph{Results}

We plot the online code results on the top of Figure \ref{fig:mdl}. The overall codelength shows a similar trend to edge probing: Most of the reduction in feature codelength is achieved with fewer than 100M words. MDL for syntactic features decreases even sooner. Results for Winograd Coref.~are idiosyncratic, probably due to the failure of the probes to learn the task.

The changes in model codelength and data codelength are shown on the bottom of Figure \ref{fig:mdl}. We compute the data codelength following \citet{voita2020informationtheoretic} using the training set loss of a classifier trained on the entire training set, and the model codelength is the total codelength minus the data codelength. The monotonically decreasing data codelength simply reflects the fact that the more data rich RoBERTa models have smaller loss. When it comes to the model codelength, however, we generally observe the global minimum for the randomly initialized models (at ``None''). This is expected, and simply reflects the fact that the decoder can barely extract any feature information from the random representations (i.e.~the probe can barely learn to recognize these features even given the full training set). On many tasks, the model codelength starts to decrease when the pretraining data volume is large enough, suggesting that large-scale pretraining may increase the data regularity of the feature information in the representations, making them more accessible to a downstream classifier. However, the decreasing trend is not consistent among all tasks, and therefore more evidence needs to be collected before we reach any conclusions about feature accessibility.

\section{Unsupervised Grammaticality Judgement}
\begin{figure*}[ht!]
    \centering
    \includegraphics[width=\textwidth]{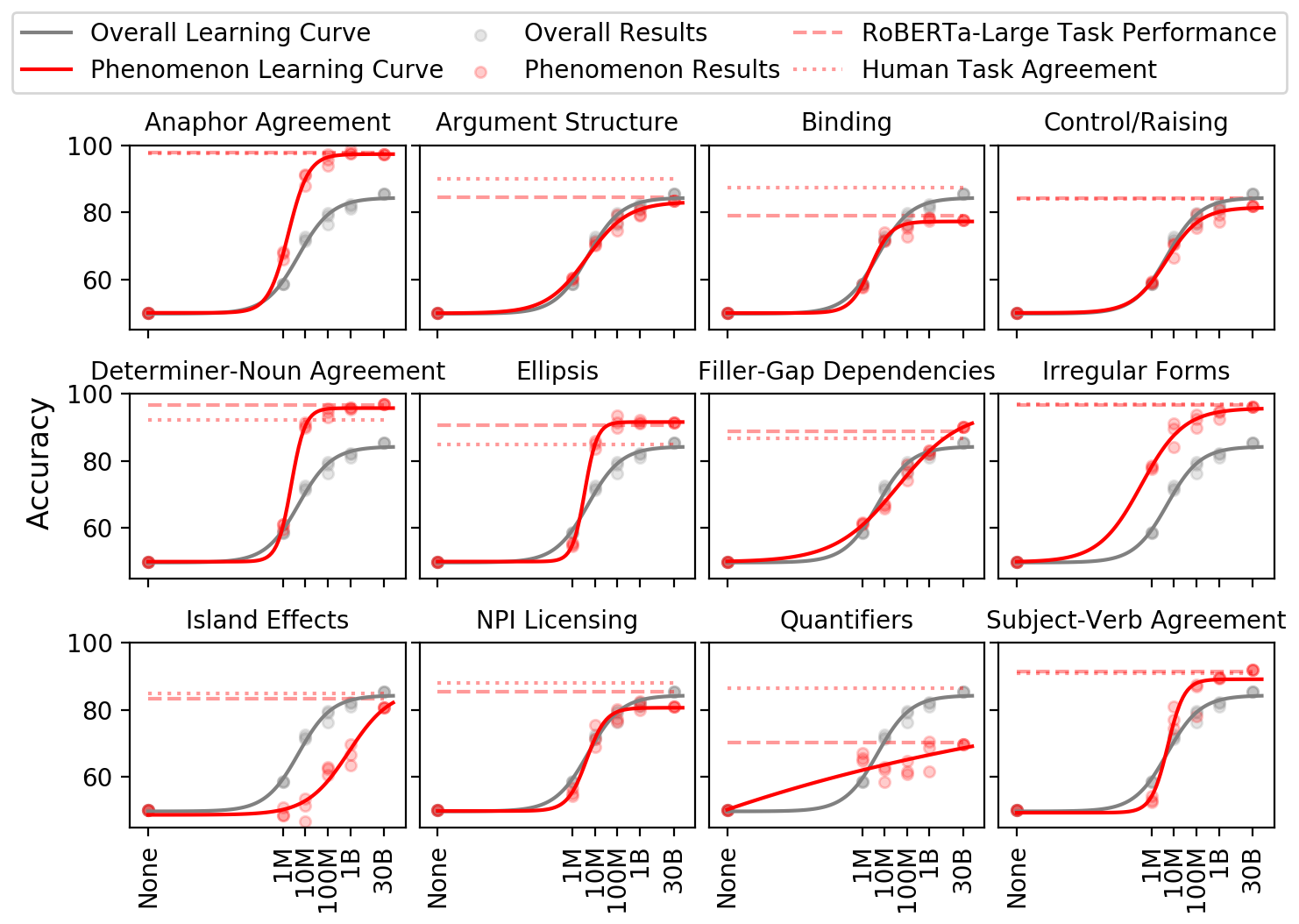}
    \caption{BLiMP results by category. BLiMP has 67 tasks which belongs to 12 linguistic phenomena. For each task the objective is to predict the more grammatically acceptable sentence of a minimal pair in an unsupervised setting. For context, we also plot human agreement with BLiMP reported by \citet{warstadt2020blimp} and RoBERTa\subtxt{LARGE} performance reported by \citet{salazar-etal-2020-masked}.}
    \label{fig:blimp}
\end{figure*}

We use the BLiMP benchmark \cite{warstadt2020blimp} to test models' knowledge of individual grammatical phenomena in English. BLiMP is a challenge set of 67 tasks, each containing 1000 minimal pairs of sentences that highlight a particular morphological, syntactic, or semantic phenomena. Minimal pairs in BLiMP consist of two sentences that differ only by a single edit, but contrast in grammatical acceptability. BLiMP is designed for unsupervised evaluation of language models using a forced choice acceptability judgment task: A language model classifies a minimal pair correctly if it assigns a higher likelihood to the acceptable sentence. We follow the MLM scoring method of \citet{salazar-etal-2020-masked} to compare candidates.

\paragraph{Results}

We plot learning curves for BLiMP in Figure \ref{fig:blimp}. \citeauthor{warstadt2020blimp} organize the 67 tasks in BLiMP into 12 categories based on the phenomena tested and for each category we plot the average accuracy for the tasks in the category. We do not normalize results in this plot. For the no-data baseline, we plot chance accuracy of 50\% rather than making empirical measurements from random RoBERTa models.

We find the greatest improvement in overall BLiMP performance between 1M and 100M words of pretraining data. With 100M words, sensitivity to contrasts in acceptability overall is within 9 accuracy points of humans, and improve only 6 points with additional data. This shows that substantial knowledge of many grammatical phenomena can be acquired from 100M words of raw text.

We also observe significant variation in how much data is needed to learn different phenomena. We see the steepest learning curves on agreement phenomena, with nearly all improvements occurring between 1M and 10M words. For phenomena involving \emph{wh}-dependencies, i.e.~filler-gap dependencies and island effects, we observe shallow and delayed learning curves with 90\% of possible improvements occurring between 1M and 100M words. These differences can most likely be ascribed to two factors: First, agreement phenomena tend to involve more local dependencies, while wh-dependencies tend to be long-distance. Second, agreement phenomena are highly frequent, with a large proportion of sentences containing multiple instances of determiner-noun and subject-verb agreement, while wh-dependencies are comparatively rare. Finally, we observe that the phenomena tested in the quantifiers category are never effectively learned, even by RoBERTa\subtxt{BASE}. These phenomena include subtle semantic contrasts---for example \emph{Nobody ate \{more than, *at least\} two cookies}---which may involve difficult-to-learn pragmatic knowledge \cite{cohen2014superlative}.

\section{Finetuning on NLU Tasks}

\begin{figure*}[ht!]
    \centering
    \includegraphics[width=\textwidth]{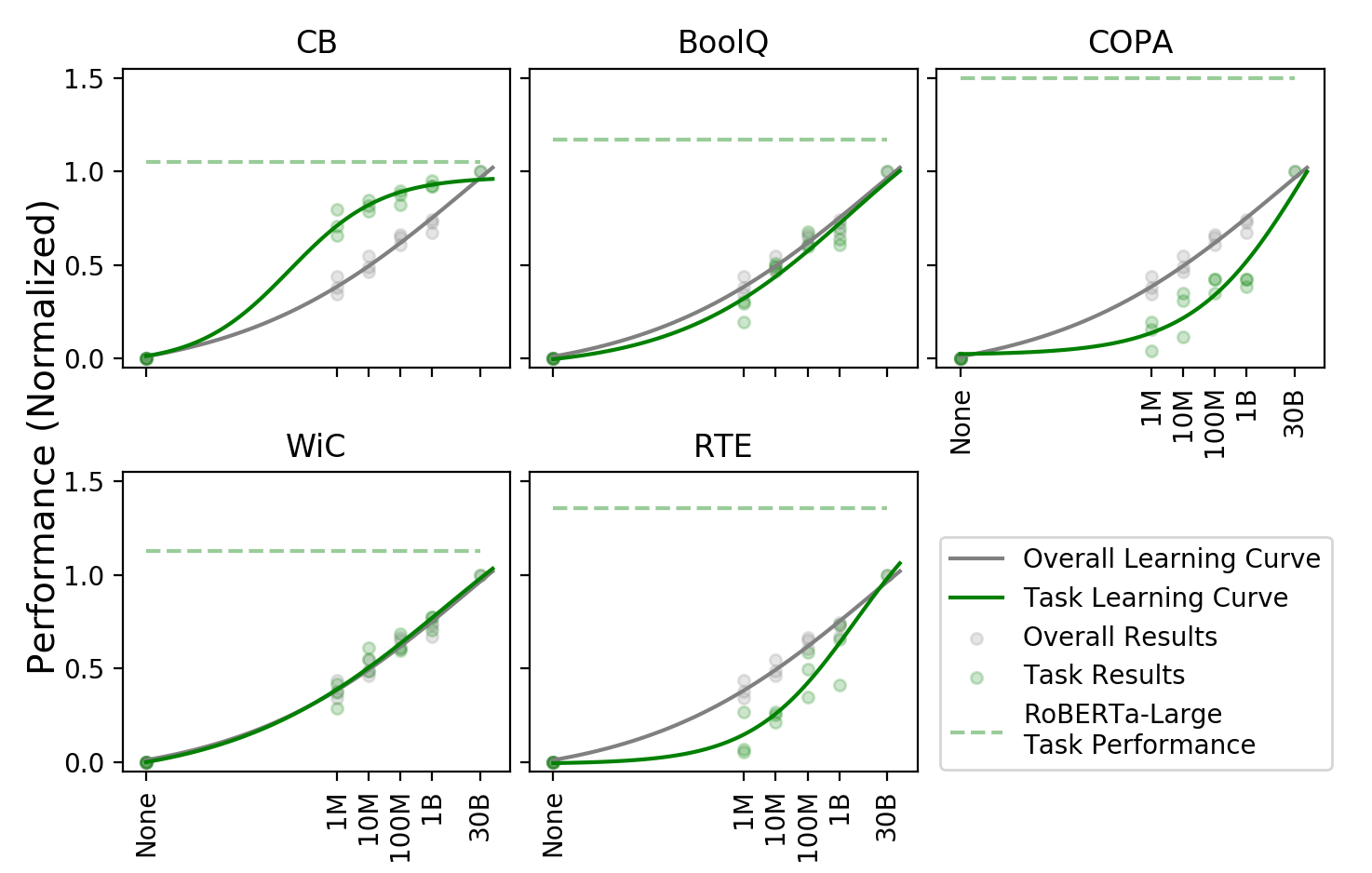}
    \caption{SuperGLUE results. The metric for BoolQ, COPA, WiC, RTE is accuracy, and for CB it is the average of accuracy and F1 score. For context, we plot RoBERTa\subtxt{LARGE} performance reported at \href{https://github.com/pytorch/fairseq/tree/master/examples/roberta}{\url{https://github.com/pytorch/fairseq/tree/master/examples/roberta}}.}
    \label{fig:superglue}
\end{figure*}

SuperGLUE is a benchmark suite of eight classification-based language-understanding tasks \citep{wang2019superglue}. We test each MiniBERTa on five SuperGLUE tasks that we expect to see significant variation at these scales.\footnote{Task Data source:
CB from \citet{demarneffe2019commitmentbank}, BoolQ from \citet{clark2019boolq}, COPA from \citet{roemmele2011choice}, WiC from \citet{pilehvar2018wic,miller1995wordnet, schuler2005verbnet}, RTE from \citet{dagan2006pascal, bar2006second, giampiccolo2007third, bentivogli2009fifth}
} The hyperparameter search range used for each task is described in the appendix.


\paragraph{Results}
We plot the results on SuperGLUE in Figure \ref{fig:superglue}. Improvements in SuperGLUE performance require a relatively large volume of pretraining data. For most tasks, the point of fastest improvement in our interpolated curve occurs with more than 1B words. None of the tasks (with the possible exception of CommitmentBank) show any significant sign of saturation at 30B words. This suggests that some key NLU skills are not learnt with fewer than billions of words, and that models are likely to continue improving on these tasks given 10 to 100 times more pretraining data.

\section{Discussion}

Having established the learning curves for each of these probing methods individually, we can begin to draw some broader conclusions about how increasing pretraining data affects Transformer MLMs. Figure \ref{fig:overall} plots the overall learning curves for these four methods together. The most striking result is that improvements in NLU task performance require far more data than improvements in representations of linguistic features as measured by these methods. Classifier probing, MDL probing, and acceptability judgment performance all improve rapidly between 1M and 10M words and show little improvement beyond 100M words. By contrast, performance on the NLU tasks in SuperGLUE appears to improve most rapidly with over 1B words and likely continues improving at much larger data scales.

This implies that at least some of the skills that RoBERTa uses to solve typical NLU  tasks require billions of words to be acquired. It is likely then that the features being tested by the edge probing suite and BLiMP are not the key skills implicated in improvements in NLU performance at these large scales. While edge probing features such as dependency and semantic role are undoubtedly crucial to solving NLU tasks, a model that can extract and encode a large proportion of these features (e.g. the 100M word models) may still perform poorly on SuperGLUE.

Commonsense knowledge may play a large role in explaining SuperGLUE performance. This hypothesis is backed up by results from the Winograd edge-probing task, which suggest that relatively little commonsense knowledge can be learned with fewer than 1B words. The notion that commonsense knowledge takes more data to learn is not surprising: Intuitively, humans mainly acquire commonsense knowledge through non-linguistic information, and so a model learning from text without grounding should require far more data than a human is exposed to in order to acquire comparable knowledge. However, as our experiments focus mainly on linguistic knowledge, additional work is needed to give a more complete picture of the acquisition of commonsense knowledge. 

Another possible explanation of the delay in the rise of the SuperGLUE curve is that being able to encode certain features does not imply being  able to use them to solve practical tasks. In other words, even if a RoBERTa model pretrained on 10M--100M words is already able to represent the linguistic feature we target, it is not guaranteed that it is able to use them in downstream tasks. This corresponds to the finding of \citet{warstadt-etal-2020-learning} that RoBERTa can learn to reliably extract many linguistic features with little pretraining data, but that it takes orders of magnitude more pretraining data for those features to be used preferentially when generalizing. Therefore, it may be a promising research direction to develop methods to efficiently pretrain and fine-tune NLP models to make better use of the linguistic features they already recognized. 


In light of \citeauthor{warstadt-etal-2020-learning}'s \citeyear{warstadt-etal-2020-learning} findings we had initially hypothesized that feature accessibility as measured by MDL might show a shallower or later learning curve than other probing methods.\footnote{\citeauthor{warstadt-etal-2020-learning}'s experiments are quite different to ours. They measure RoBERTa's preference for linguistic features over surface features during fine-tuning on ambiguous classification tasks. This requires strong inductive biases, which may not correspond straightforwardly to MDL.} This hypothesis is not supported by our findings: Figure \ref{fig:overall} shows no obvious difference between the classifier probing curve and the MDL probing curve. However, this does not prove that the accessibility of linguistic features does not improve with massive pretraining sets, nor does it prove that the information about a feature and its accessibility improve at the same rate. 

While those conclusions may turn out to be correct, another possibility is that the setting and methods we adopt fail to adequately differentiate between feature information and accessibility. The bottom of Figure \ref{fig:mdl} shows that for most tasks the data codelength has a much larger variance across pretraining volumes than the model codelength, and thus the change in overall codelength predominantly reflects the decrease in the loss of a classifier trained on the full training set. Therefore, it is not surprising that the MDL curve resembles that of classifier probing. However, comparing model codelengths alone does not reliably reveal feature accessibility either, since model codelength is not optimized individually but as a part of the overall codelength. New probing methods related to MDL address different aspects of these problems \cite{whitney2020evaluating,pimentel2020pareto} and may yield different conclusions.


\section{Related Work}

Probing neural network representations has been an active area of research in recent years \citep{rogers2020primer,Belinkov2019AnalysisMI}. With the advent of large pretrained Transformers like BERT \cite{devlin2019bert}, numerous papers have used classifier probes methods to attempt to locate linguistic features in learned representations with striking positive results \cite{tenney2019you,hewitt2019structural}. However, another thread has found problems with many probing methods: Classifier probes can learn too much from training data \cite{hewitt2019designing} and can fail to distinguish between features that are extractable and features that are actually used \cite{voita2020informationtheoretic,pimentel2020information,elazar2020amnesic}. Moreover, it is advisable to look to a variety of probing methods, as different probing methods often yield contradictory results \cite{warstadt2019investigating}. 

There have also been a few earlier studies investigating the relationship between pretraining data volume and linguistic knowledge in language models. Studies of unsupervised acceptability judgments find fairly consistent evidence of rapid improvements in linguistic knowledge up to about 10M words of pretraining data, after which improvements slow down for most phenomena. \citet{vanschijndel2019quantity} find large improvements in knowledge of subject-verb agreement and reflexive binding up to 10M words, and few improvements between 10M and 80M words. \citet{hu-etal-2020-systematic} find that GPT-2 trained on 42M words performs roughly as well on a syntax benchmark as a similar model trained on 100 times that amount. Other studies have investigate how one model's linguistic knowledge changes during the training process, as a function of the number of updates \cite{saphra2019understanding,chiang2020pretrained}.

\citet{JMLR:v21:20-074} also investigate how performance on SuperGLUE (and other downstream tasks) improves with pretraining dataset sizes between about 8M and 34B words. In contrast to our findings, they find that models with around 500M words of pretraining data can perform similarly on downstream tasks to models with 34B words. This discrepancy may arise from several factors. First, the architecture and pretraining for their T5 model is not identical to RoBERTa's or the MiniBERTas'. Second, they pretrain their models for a fixed number of iterations (totaling 34B tokens), whereas the miniBERTas were trained with early stopping. Nonetheless, this result suggests that the number of unique tokens might matter less than the number of iterations, within reasonable limits.

There is also some recent work that investigates the effect of pretraining data size of other languages. \citet{micheli-etal-2020-importance} pretrain BERT-based language models on 10MB, 100MB, 500MB, 1000MB, 2000MB, and 4000MB of French text and test them on a question answering task. They find that the French MLM pretrained on 100MB of raw text has similar performance to the ones pretrained on larger datasets on the task, and that corpus-specific self-supervised learning does not make a significant difference. \citet{martin-etal-2020-camembert} also show that French MLMs can already learn a lot from small-scale pretraining.

\section{Conclusion}

We track the ability of language models to acquire representations of linguistic features as a function of the amount of pretraining data. We use a variety of probing methods, from which we determine that linguistic features are mostly learnable with 100M words of data, while NLU task performance requires far more data.

Our results do not explain what causes NLU task performance to improve with large quantities of data. To answer these questions, one can use causal probing methods like amnesic probing \cite{elazar2020amnesic}, in which features are removed from a representation. We would also like to understand the differences between learning curves for various linguistic features, for instance through the lens of the hypothesis that some features acquired earlier on play a role in bootstrapping knowledge of other features? Finally, our results show that to the extent that Transformer LMs like RoBERTa even approach human language understanding, they require far more data than humans to do so. Extending this investigation to other pretraining settings and study different model architectures, pretraining tasks, and pretraining data domains---including ones that more closely resemble human learners---could help indicate promising directions for closing this gap.

\section*{Acknowledgments}
We thank Haokun Liu and Ian Tenney for providing technical support on the edge probing experiment, and Elena Voita for support with MDL. This project has benefited from financial support to SB by Eric and Wendy Schmidt (made by recommendation of the Schmidt Futures program), by Samsung Research (under the project \textit{Improving Deep Learning using Latent Structure}), by Intuit, Inc., and in-kind support by the NYU High-Performance Computing Center. This material is based upon work supported by the National Science Foundation under Grant Nos. 1850208 and 1922658. Any opinions, findings, and conclusions or recommendations expressed in this material are those of the author(s) and do not necessarily reflect the views of the National Science Foundation. 

\bibliography{anthology,eacl2021}

\begin{thebibliography}{52}
\expandafter\ifx\csname natexlab\endcsname\relax\def\natexlab#1{#1}\fi

\bibitem[{Adi et~al.(2017)Adi, Kermany, Belinkov, Lavi, and
  Goldberg}]{adi2017fine}
Yossi Adi, Einat Kermany, Yonatan Belinkov, Ofer Lavi, and Yoav Goldberg. 2017.
\newblock Fine-grained analysis of sentence embeddings using auxiliary
  prediction tasks.
\newblock In \emph{Proceedings of ICLR Conference Track. Toulon, France.}

\bibitem[{Bar~Haim et~al.(2006)Bar~Haim, Dagan, Dolan, Ferro, Giampiccolo,
  Magnini, and Szpektor}]{bar2006second}
Roy Bar~Haim, Ido Dagan, Bill Dolan, Lisa Ferro, Danilo Giampiccolo, Bernardo
  Magnini, and Idan Szpektor. 2006.
\newblock \href {http://u.cs.biu.ac.il/~nlp/RTE2/Proceedings/01.pdf} {The
  second {PASCAL} recognising textual entailment challenge}.
\newblock In \emph{Proceedings of the Second {PASCAL} Challenges Workshop on
  Recognising Textual Entailment}.

\bibitem[{Belinkov and Glass(2019)}]{Belinkov2019AnalysisMI}
Yonatan Belinkov and James~R. Glass. 2019.
\newblock Analysis methods in neural language processing: A survey.
\newblock \emph{Transactions of the Association for Computational Linguistics},
  7:49--72.

\bibitem[{Bentivogli et~al.(2009)Bentivogli, Dagan, Dang, Giampiccolo, and
  Magnini}]{bentivogli2009fifth}
Luisa Bentivogli, Ido Dagan, Hoa~Trang Dang, Danilo Giampiccolo, and Bernardo
  Magnini. 2009.
\newblock \href
  {http://citeseerx.ist.psu.edu/viewdoc/summary?doi=10.1.1.232.1231} {The fifth
  {PASCAL} recognizing textual entailment challenge}.
\newblock In \emph{Textual Analysis Conference (TAC)}.

\bibitem[{Brown et~al.(2020)Brown, Mann, Ryder, Subbiah, Kaplan, Dhariwal,
  Neelakantan, Shyam, Sastry, Askell, Agarwal, Herbert-Voss, Krueger, Henighan,
  Child, Ramesh, Ziegler, Wu, Winter, Hesse, Chen, Sigler, Litwin, Gray, Chess,
  Clark, Berner, McCandlish, Radford, Sutskever, and Amodei}]{brown2020gpt3}
Tom~B. Brown, Benjamin Mann, Nick Ryder, Melanie Subbiah, Jared Kaplan,
  Prafulla Dhariwal, Arvind Neelakantan, Pranav Shyam, Girish Sastry, Amanda
  Askell, Sandhini Agarwal, Ariel Herbert-Voss, Gretchen Krueger, Tom Henighan,
  Rewon Child, Aditya Ramesh, Daniel~M. Ziegler, Jeffrey Wu, Clemens Winter,
  Christopher Hesse, Mark Chen, Eric Sigler, Mateusz Litwin, Scott Gray,
  Benjamin Chess, Jack Clark, Christopher Berner, Sam McCandlish, Alec Radford,
  Ilya Sutskever, and Dario Amodei. 2020.
\newblock Language models are few-shot learners.
\newblock \emph{arXiv preprint 2005.14165}.

\bibitem[{Chiang et~al.(2020)Chiang, Huang, and Lee}]{chiang2020pretrained}
David~C Chiang, Sung-Feng Huang, and Hung-yi Lee. 2020.
\newblock Pretrained language model embryology: The birth of {ALBERT}.
\newblock \emph{arXiv preprint arXiv:2010.02480}.

\bibitem[{Clark et~al.(2019)Clark, Lee, Chang, Kwiatkowski, Collins, and
  Toutanova}]{clark2019boolq}
Christopher Clark, Kenton Lee, Ming-Wei Chang, Tom Kwiatkowski, Michael
  Collins, and Kristina Toutanova. 2019.
\newblock Boolq: Exploring the surprising difficulty of natural yes/no
  questions.
\newblock In \emph{Proceedings of the 2019 Conference of the North American
  Chapter of the Association for Computational Linguistics: Human Language
  Technologies, Volume 1 (Long and Short Papers)}, pages 2924--2936.

\bibitem[{Cohen and Krifka(2014)}]{cohen2014superlative}
Ariel Cohen and Manfred Krifka. 2014.
\newblock Superlative quantifiers and meta-speech acts.
\newblock \emph{Linguistics and Philosophy}, 37(1):41--90.

\bibitem[{Dagan et~al.(2006)Dagan, Glickman, and Magnini}]{dagan2006pascal}
Ido Dagan, Oren Glickman, and Bernardo Magnini. 2006.
\newblock \href {https://link.springer.com/chapter/10.1007/11736790_9} {The
  {PASCAL} recognising textual entailment challenge}.
\newblock In \emph{Machine Learning Challenges. Evaluating Predictive
  Uncertainty, Visual Object Classification, and Recognising Textual
  Entailment}. Springer.

\bibitem[{De~Marneffe et~al.(2019)De~Marneffe, Simons, and
  Tonhauser}]{demarneffe2019commitmentbank}
Marie-Catherine De~Marneffe, Mandy Simons, and Judith Tonhauser. 2019.
\newblock The commitmentbank: Investigating projection in naturally occurring
  discourse.
\newblock In \emph{proceedings of Sinn und Bedeutung}, volume~23, pages
  107--124.

\bibitem[{Devlin et~al.(2019)Devlin, Chang, Lee, and
  Toutanova}]{devlin2019bert}
Jacob Devlin, Ming-Wei Chang, Kenton Lee, and Kristina Toutanova. 2019.
\newblock {BERT}: Pre-training of deep bidirectional transformers for language
  understanding.
\newblock In \emph{Proceedings of the 2019 Conference of the North American
  Chapter of the Association for Computational Linguistics: Human Language
  Technologies, Volume 1 (Long and Short Papers)}, pages 4171--4186.

\bibitem[{Elazar et~al.(2020)Elazar, Ravfogel, Jacovi, and
  Goldberg}]{elazar2020amnesic}
Yanai Elazar, Shauli Ravfogel, Alon Jacovi, and Yoav Goldberg. 2020.
\newblock When {BERT} forgets how to {POS}: {A}mnesic probing of linguistic
  properties and {MLM} predictions.
\newblock \emph{arXiv preprint 2006.00995}.

\bibitem[{Ettinger(2020)}]{ettinger2020bert}
Allyson Ettinger. 2020.
\newblock \href {https://doi.org/10.1162/tacl\_a\_00298} {What {BERT} is not:
  Lessons from a new suite of psycholinguistic diagnostics for language
  models}.
\newblock \emph{Transactions of the Association for Computational Linguistics},
  8:34--48.

\bibitem[{Ettinger et~al.(2016)Ettinger, Elgohary, and
  Resnik}]{ettinger2016probing}
Allyson Ettinger, Ahmed Elgohary, and Philip Resnik. 2016.
\newblock Probing for semantic evidence of composition by means of simple
  classification tasks.
\newblock In \emph{Proceedings of the 1st Workshop on Evaluating Vector-Space
  Representations for NLP}, pages 134--139.

\bibitem[{Giampiccolo et~al.(2007)Giampiccolo, Magnini, Dagan, and
  Dolan}]{giampiccolo2007third}
Danilo Giampiccolo, Bernardo Magnini, Ido Dagan, and Bill Dolan. 2007.
\newblock The third {PASCAL} recognizing textual entailment challenge.
\newblock In \emph{Proceedings of the {ACL-PASCAL} Workshop on Textual
  Entailment and Paraphrasing}. Association for Computational Linguistics.

\bibitem[{Hart and Risley(1992)}]{hart1992parenting}
Betty Hart and Todd~R. Risley. 1992.
\newblock American parenting of language-learning children: Persisting
  differences in family-child interactions observed in natural home
  environments.
\newblock \emph{Developmental Psychology}, 28(6):1096.

\bibitem[{Heathcote et~al.(2000)Heathcote, Brown, and
  Mewhort}]{heathcote2000power}
Andrew Heathcote, Scott Brown, and Douglas~JK Mewhort. 2000.
\newblock The power law repealed: The case for an exponential law of practice.
\newblock \emph{Psychonomic bulletin \& review}, 7(2):185--207.

\bibitem[{Heim and Kratzer(1998)}]{heim1998semantics}
Irene Heim and Angelika Kratzer. 1998.
\newblock \emph{Semantics in generative grammar}.
\newblock Blackwell Oxford.

\bibitem[{Hendrickx et~al.(2010)Hendrickx, Kim, Kozareva, Nakov,
  {\'O}~S{\'e}aghdha, Pad{\'o}, Pennacchiotti, Romano, and
  Szpakowicz}]{hendrickx-etal-2010-semeval}
Iris Hendrickx, Su~Nam Kim, Zornitsa Kozareva, Preslav Nakov, Diarmuid
  {\'O}~S{\'e}aghdha, Sebastian Pad{\'o}, Marco Pennacchiotti, Lorenza Romano,
  and Stan Szpakowicz. 2010.
\newblock \href {https://www.aclweb.org/anthology/S10-1006} {{S}em{E}val-2010
  task 8: Multi-way classification of semantic relations between pairs of
  nominals}.
\newblock In \emph{Proceedings of the 5th International Workshop on Semantic
  Evaluation}, pages 33--38, Uppsala, Sweden. Association for Computational
  Linguistics.

\bibitem[{Hewitt and Liang(2019)}]{hewitt2019designing}
John Hewitt and Percy Liang. 2019.
\newblock Designing and interpreting probes with control tasks.
\newblock In \emph{Conference on Empirical Methods in Natural Language
  Processing}. Association for Computational Linguistics.

\bibitem[{Hewitt and Manning(2019)}]{hewitt2019structural}
John Hewitt and Christopher~D Manning. 2019.
\newblock A structural probe for finding syntax in word representations.
\newblock In \emph{Proceedings of the 2019 Conference of the North American
  Chapter of the Association for Computational Linguistics: Human Language
  Technologies, Volume 1 (Long and Short Papers)}, pages 4129--4138.

\bibitem[{Hu et~al.(2020)Hu, Gauthier, Qian, Wilcox, and
  Levy}]{hu-etal-2020-systematic}
Jennifer Hu, Jon Gauthier, Peng Qian, Ethan Wilcox, and Roger Levy. 2020.
\newblock \href {https://www.aclweb.org/anthology/2020.acl-main.158} {A
  systematic assessment of syntactic generalization in neural language models}.
\newblock In \emph{Proceedings of the 58th Annual Meeting of the Association
  for Computational Linguistics}, pages 1725--1744, Online. Association for
  Computational Linguistics.

\bibitem[{Kingma and Ba(2014)}]{kingma2014adam}
Diederik~P. Kingma and Jimmy Ba. 2014.
\newblock Adam: A method for stochastic optimization.
\newblock In \emph{Proceedings of the 3rd International Conference on Learning
  Representations}.

\bibitem[{Liu et~al.(2019)Liu, Ott, Goyal, Du, Joshi, Chen, Levy, Lewis,
  Zettlemoyer, and Stoyanov}]{liu2019roberta}
Yinhan Liu, Myle Ott, Naman Goyal, Jingfei Du, Mandar Joshi, Danqi Chen, Omer
  Levy, Mike Lewis, Luke Zettlemoyer, and Veselin Stoyanov. 2019.
\newblock Roberta: A robustly optimized bert pretraining approach.
\newblock \emph{arXiv preprint arXiv:1907.11692}.

\bibitem[{Martin et~al.(2020)Martin, Muller, Ortiz~Su{\'a}rez, Dupont, Romary,
  de~la Clergerie, Seddah, and Sagot}]{martin-etal-2020-camembert}
Louis Martin, Benjamin Muller, Pedro~Javier Ortiz~Su{\'a}rez, Yoann Dupont,
  Laurent Romary, {\'E}ric de~la Clergerie, Djam{\'e} Seddah, and Beno{\^\i}t
  Sagot. 2020.
\newblock \href {https://doi.org/10.18653/v1/2020.acl-main.645} {{C}amem{BERT}:
  a tasty {F}rench language model}.
\newblock In \emph{Proceedings of the 58th Annual Meeting of the Association
  for Computational Linguistics}, pages 7203--7219, Online. Association for
  Computational Linguistics.

\bibitem[{Micheli et~al.(2020)Micheli, d{'}Hoffschmidt, and
  Fleuret}]{micheli-etal-2020-importance}
Vincent Micheli, Martin d{'}Hoffschmidt, and Fran{\c{c}}ois Fleuret. 2020.
\newblock \href {https://www.aclweb.org/anthology/2020.emnlp-main.632} {On the
  importance of pre-training data volume for compact language models}.
\newblock In \emph{Proceedings of the 2020 Conference on Empirical Methods in
  Natural Language Processing (EMNLP)}, pages 7853--7858, Online. Association
  for Computational Linguistics.

\bibitem[{Miller(1995)}]{miller1995wordnet}
George~A Miller. 1995.
\newblock \href {https://www.aclweb.org/anthology/H94-1111} {{WordNet}: a
  lexical database for english}.
\newblock \emph{Communications of the ACM}.

\bibitem[{Pilehvar and Camacho-Collados(2019)}]{pilehvar2018wic}
Mohammad~Taher Pilehvar and Jose Camacho-Collados. 2019.
\newblock \href {https://arxiv.org/abs/1808.09121} {{WiC}: The word-in-context
  dataset for evaluating context-sensitive meaning representations}.
\newblock In \emph{Proceedings of the Conference of the North American Chapter
  of the Association for Computational Linguistics: Human Language Technologies
  (NAACL-HLT)}. Association for Computational Linguistics.

\bibitem[{Pimentel et~al.(2020{\natexlab{a}})Pimentel, Saphra, Williams, and
  Cotterell}]{pimentel2020pareto}
Tiago Pimentel, Naomi Saphra, Adina Williams, and Ryan Cotterell.
  2020{\natexlab{a}}.
\newblock Pareto probing: Trading off accuracy for complexity.
\newblock \emph{arXiv preprint arXiv:2010.02180}.

\bibitem[{Pimentel et~al.(2020{\natexlab{b}})Pimentel, Valvoda, Hall~Maudslay,
  Zmigrod, Williams, and Cotterell}]{pimentel2020information}
Tiago Pimentel, Josef Valvoda, Rowan Hall~Maudslay, Ran Zmigrod, Adina
  Williams, and Ryan Cotterell. 2020{\natexlab{b}}.
\newblock \href {https://doi.org/10.18653/v1/2020.acl-main.420}
  {Information-theoretic probing for linguistic structure}.
\newblock In \emph{Proceedings of the 58th Annual Meeting of the Association
  for Computational Linguistics}, pages 4609--4622, Online. Association for
  Computational Linguistics.

\bibitem[{Pruksachatkun et~al.(2020)Pruksachatkun, Phang, Liu, Htut, Zhang,
  Pang, Vania, Kann, and Bowman}]{pruksachatkun-etal-2020-intermediate}
Yada Pruksachatkun, Jason Phang, Haokun Liu, Phu~Mon Htut, Xiaoyi Zhang,
  Richard~Yuanzhe Pang, Clara Vania, Katharina Kann, and Samuel~R. Bowman.
  2020.
\newblock \href {https://doi.org/10.18653/v1/2020.acl-main.467}
  {Intermediate-task transfer learning with pretrained language models: When
  and why does it work?}
\newblock In \emph{Proceedings of the 58th Annual Meeting of the Association
  for Computational Linguistics}, pages 5231--5247, Online. Association for
  Computational Linguistics.

\bibitem[{Raffel et~al.(2020)Raffel, Shazeer, Roberts, Lee, Narang, Matena,
  Zhou, Li, and Liu}]{JMLR:v21:20-074}
Colin Raffel, Noam Shazeer, Adam Roberts, Katherine Lee, Sharan Narang, Michael
  Matena, Yanqi Zhou, Wei Li, and Peter~J. Liu. 2020.
\newblock \href {http://jmlr.org/papers/v21/20-074.html} {Exploring the limits
  of transfer learning with a unified text-to-text transformer}.
\newblock \emph{Journal of Machine Learning Research}, 21(140):1--67.

\bibitem[{Rahman and Ng(2012)}]{rahman-ng-2012-resolving}
Altaf Rahman and Vincent Ng. 2012.
\newblock \href {https://www.aclweb.org/anthology/D12-1071} {Resolving complex
  cases of definite pronouns: The {W}inograd schema challenge}.
\newblock In \emph{Proceedings of the 2012 Joint Conference on Empirical
  Methods in Natural Language Processing and Computational Natural Language
  Learning}, pages 777--789, Jeju Island, Korea. Association for Computational
  Linguistics.

\bibitem[{Roemmele et~al.(2011)Roemmele, Bejan, and
  Gordon}]{roemmele2011choice}
Melissa Roemmele, Cosmin~Adrian Bejan, and Andrew~S. Gordon. 2011.
\newblock Choice of plausible alternatives: An evaluation of commonsense causal
  reasoning.
\newblock In \emph{2011 AAAI Spring Symposium Series}.

\bibitem[{Rogers et~al.(2020)Rogers, Kovaleva, and
  Rumshisky}]{rogers2020primer}
Anna Rogers, Olga Kovaleva, and Anna Rumshisky. 2020.
\newblock A primer in {BERT}ology: What we know about how {BERT} works.
\newblock In \emph{Findings of EMNLP}.

\bibitem[{Rudinger et~al.(2018)Rudinger, Teichert, Culkin, Zhang, and
  Van~Durme}]{rudinger-etal-2018-neural-davidsonian}
Rachel Rudinger, Adam Teichert, Ryan Culkin, Sheng Zhang, and Benjamin
  Van~Durme. 2018.
\newblock \href {https://doi.org/10.18653/v1/D18-1114} {Neural-{D}avidsonian
  semantic proto-role labeling}.
\newblock In \emph{Proceedings of the 2018 Conference on Empirical Methods in
  Natural Language Processing}, pages 944--955, Brussels, Belgium. Association
  for Computational Linguistics.

\bibitem[{Salazar et~al.(2020)Salazar, Liang, Nguyen, and
  Kirchhoff}]{salazar-etal-2020-masked}
Julian Salazar, Davis Liang, Toan~Q. Nguyen, and Katrin Kirchhoff. 2020.
\newblock \href {https://doi.org/10.18653/v1/2020.acl-main.240} {Masked
  language model scoring}.
\newblock In \emph{Proceedings of the 58th Annual Meeting of the Association
  for Computational Linguistics}, pages 2699--2712, Online. Association for
  Computational Linguistics.

\bibitem[{Saphra and Lopez(2019)}]{saphra2019understanding}
Naomi Saphra and Adam Lopez. 2019.
\newblock \href {https://doi.org/10.18653/v1/N19-1329} {Understanding learning
  dynamics of language models with {SVCCA}}.
\newblock In \emph{Proceedings of the 2019 Conference of the North {A}merican
  Chapter of the Association for Computational Linguistics: Human Language
  Technologies, Volume 1 (Long and Short Papers)}, pages 3257--3267,
  Minneapolis, Minnesota. Association for Computational Linguistics.

\bibitem[{van Schijndel et~al.(2019)van Schijndel, Mueller, and
  Linzen}]{vanschijndel2019quantity}
Marten van Schijndel, Aaron Mueller, and Tal Linzen. 2019.
\newblock \href {https://doi.org/10.18653/v1/D19-1592} {Quantity doesn{'}t buy
  quality syntax with neural language models}.
\newblock In \emph{Proceedings of the 2019 Conference on Empirical Methods in
  Natural Language Processing and the 9th International Joint Conference on
  Natural Language Processing (EMNLP-IJCNLP)}, pages 5831--5837, Hong Kong,
  China. Association for Computational Linguistics.

\bibitem[{Schuler(2005)}]{schuler2005verbnet}
Karin~Kipper Schuler. 2005.
\newblock \href {http://verbs.colorado.edu/~kipper/Papers/dissertation.pdf}
  {\emph{Verbnet: A Broad-coverage, Comprehensive Verb Lexicon}}.
\newblock Ph.D. thesis, University of Pennsylvania.

\bibitem[{Silveira et~al.(2014)Silveira, Dozat, de~Marneffe, Bowman, Connor,
  Bauer, and Manning}]{silveira-etal-2014-gold}
Natalia Silveira, Timothy Dozat, Marie-Catherine de~Marneffe, Samuel Bowman,
  Miriam Connor, John Bauer, and Chris Manning. 2014.
\newblock \href
  {http://www.lrec-conf.org/proceedings/lrec2014/pdf/1089_Paper.pdf} {A gold
  standard dependency corpus for {E}nglish}.
\newblock In \emph{Proceedings of the Ninth International Conference on
  Language Resources and Evaluation ({LREC}-2014)}, pages 2897--2904,
  Reykjavik, Iceland. European Languages Resources Association (ELRA).

\bibitem[{Teichert et~al.(2017)Teichert, Poliak, Durme, and
  Gormley}]{AAAI1714997}
Adam Teichert, Adam Poliak, Benjamin~Van Durme, and Matthew Gormley. 2017.
\newblock \href
  {https://www.aaai.org/ocs/index.php/AAAI/AAAI17/paper/view/14997} {Semantic
  proto-role labeling}.
\newblock In \emph{AAAI Conference on Artificial Intelligence}.

\bibitem[{Tenney et~al.(2019{\natexlab{a}})Tenney, Das, and
  Pavlick}]{tenney-etal-2019-bert}
Ian Tenney, Dipanjan Das, and Ellie Pavlick. 2019{\natexlab{a}}.
\newblock \href {https://doi.org/10.18653/v1/P19-1452} {{BERT} rediscovers the
  classical {NLP} pipeline}.
\newblock In \emph{Proceedings of the 57th Annual Meeting of the Association
  for Computational Linguistics}, pages 4593--4601, Florence, Italy.
  Association for Computational Linguistics.

\bibitem[{Tenney et~al.(2019{\natexlab{b}})Tenney, Xia, Chen, Wang, Poliak,
  McCoy, Kim, Van~Durme, Bowman, Das et~al.}]{tenney2019you}
Ian Tenney, Patrick Xia, Berlin Chen, Alex Wang, Adam Poliak, R.~Thomas McCoy,
  Najoung Kim, Benjamin Van~Durme, Samuel~R Bowman, Dipanjan Das, et~al.
  2019{\natexlab{b}}.
\newblock What do you learn from context? {P}robing for sentence structure in
  contextualized word representations.
\newblock In \emph{Proceedings of ICLR}.

\bibitem[{Voita and Titov(2020)}]{voita2020informationtheoretic}
Elena Voita and Ivan Titov. 2020.
\newblock Information-theoretic probing with minimum description length.
\newblock In \emph{Proceedings of the 2020 Conference on Empirical Methods in
  Natural Language Processing}, Punta Cana, Dominican Republic. Association for
  Computational Linguistics.

\bibitem[{Wang et~al.(2019)Wang, Pruksachatkun, Nangia, Singh, Michael, Hill,
  Levy, and Bowman}]{wang2019superglue}
Alex Wang, Yada Pruksachatkun, Nikita Nangia, Amanpreet Singh, Julian Michael,
  Felix Hill, Omer Levy, and Samuel~R. Bowman. 2019.
\newblock Super{GLUE}: {A} stickier benchmark for general-purpose language
  understanding systems.
\newblock In \emph{33rd Conference on Neural Information Processing Systems}.

\bibitem[{Warstadt et~al.(2019)Warstadt, Cao, Grosu, Peng, Blix, Nie, Alsop,
  Bordia, Liu, Parrish, Wang, Phang, Mohananey, Htut, Jereti\v{c}, and
  Bowman}]{warstadt2019investigating}
Alex Warstadt, Yu~Cao, Ioana Grosu, Wei Peng, Hagen Blix, Yining Nie, Anna
  Alsop, Shikha Bordia, Haokun Liu, Alicia Parrish, Sheng-Fu Wang, Jason Phang,
  Anhad Mohananey, Phu~Mon Htut, Paloma Jereti\v{c}, and Samuel~R. Bowman.
  2019.
\newblock Investigating {BERT}'s knowledge of language: {F}ive analysis methods
  with {NPI}s.
\newblock In \emph{Proceedings of EMNLP-IJCNLP}, pages 2870--2880.

\bibitem[{Warstadt et~al.(2020{\natexlab{a}})Warstadt, Parrish, Liu, Mohananey,
  Peng, Wang, and Bowman}]{warstadt2020blimp}
Alex Warstadt, Alicia Parrish, Haokun Liu, Anhad Mohananey, Wei Peng, Sheng-Fu
  Wang, and Samuel~R. Bowman. 2020{\natexlab{a}}.
\newblock \href {https://doi.org/10.1162/tacl\_a\_00321} {Blimp: The benchmark
  of linguistic minimal pairs for english}.
\newblock \emph{Transactions of the Association for Computational Linguistics},
  8:377--392.

\bibitem[{Warstadt et~al.(2020{\natexlab{b}})Warstadt, Zhang, Li, Liu, and
  Bowman}]{warstadt-etal-2020-learning}
Alex Warstadt, Yian Zhang, Haau-Sing Li, Haokun Liu, and Samuel~R Bowman.
  2020{\natexlab{b}}.
\newblock Learning which features matter: Roberta acquires a preference for
  linguistic generalizations (eventually).
\newblock In \emph{Proceedings of the 2020 Conference on Empirical Methods in
  Natural Language Processing}, Punta Cana, Dominican Republic. Association for
  Computational Linguistics.

\bibitem[{Weischedel et~al.(2013)Weischedel, Palmer, Mitchell, Hovy, Pradhan,
  Ramshaw, Xue, Taylor, Kaufman, Franchini, El-Bachouti, Belvin, and
  Houston}]{weischedel2013ontonotes}
Ralph Weischedel, Martha Palmer, Marcus Mitchell, Eduard Hovy, Sameer Pradhan,
  Lance Ramshaw, Nianwen Xue, Ann Taylor, Jeff Kaufman, Michelle Franchini,
  Mohammed El-Bachouti, Robert Belvin, and Ann Houston. 2013.
\newblock {O}nto{N}otes release 5.0 {LDC2013T19}.
\newblock Linguistic Data Consortium.

\bibitem[{White et~al.(2017)White, Rastogi, Duh, and Van~Durme}]{white2017}
Aaron~Steven White, Pushpendre Rastogi, Kevin Duh, and Benjamin Van~Durme.
  2017.
\newblock Inference is everything: Recasting semantic resources into a unified
  evaluation framework.
\newblock In \emph{Proceedings of the Eighth International Joint Conference on
  Natural Language Processing (Volume 1: Long Papers)}, pages 996--1005.

\bibitem[{Whitney et~al.(2020)Whitney, Song, Brandfonbrener, Altosaar, and
  Cho}]{whitney2020evaluating}
William~F Whitney, Min~Jae Song, David Brandfonbrener, Jaan Altosaar, and
  Kyunghyun Cho. 2020.
\newblock Evaluating representations by the complexity of learning low-loss
  predictors.
\newblock \emph{arXiv preprint arXiv:2009.07368}.

\end{thebibliography}
\bibliographystyle{acl_natbib}

\appendix

\section{Appendices}

 \begin{table*}[t]
 \centering \small
\begin{tabular}{lcccc}
\toprule
\bf Task & \bf Batch Size & \bf Learning Rate & \bf validation interval & \bf Max Epochs  \\
\midrule
 BoolQ &  \{2,4,8\} &  \{1e-6, 5e-6, 1e-5\} &  2400 & 10  \\ \midrule
 CB &  \{2,4,8\} &  \{1e-5, 5e-5, 1e-4\} &  60 & 40  \\ \midrule
 COPA &  \{16,32,64\} &  \{1e-6, 5e-6, 1e-5\} &  100 & 40  \\ \midrule
 RTE &  \{2,4,8\} &  \{5e-6, 1e-5, 5e-5\} &  1000 & 40  \\ \midrule
  WiC &  \{16,32,64\} &  \{1e-5, 5e-5, 1e-4\} &  1000 & 10  \\ 

\bottomrule
\end{tabular}
\caption{Hyperparameter search ranges for the SuperGLUE tasks. Our search ranges are largely dependent on those used in \citet{pruksachatkun-etal-2020-intermediate}. }
\end{table*}

\begin{table*}[t]
\centering\small
\resizebox{\textwidth}{!}{%
\begin{tabular}{llllllllllllll}
\begin{turn}{\ang}%
\textbf{Model}\end{turn} 
& \begin{turn}{\ang}%
\bf Overall \end{turn}
& \begin{turn}{\ang}%
\textbf{ANA. AGR} \end{turn}
& \begin{turn}{\ang}%
\textbf{ARG. STR}\end{turn}
& \begin{turn}{\ang}%
\textbf{BINDING} \end{turn}
& \begin{turn}{\ang}%
\textbf{CTRL. RAIS.}\end{turn}
& \begin{turn}{\ang}%
\textbf{D-N AGR} \end{turn}
& \begin{turn}{\ang}%
\textbf{ELLIPSIS} \end{turn}
& \begin{turn}{\ang}%
\textbf{FILLER GAP} \end{turn}
& \begin{turn}{\ang}%
\textbf{IRREGULAR} \end{turn}
& \begin{turn}{\ang}%
\textbf{ISLAND} \end{turn}
& \begin{turn}{\ang}%
\textbf{NPI} \end{turn}
& \begin{turn}{\ang}%
\textbf{QUANTIFIERS}\end{turn}
& \begin{turn}{\ang}%
\textbf{S-V AGR}\end{turn}
\\
\midrule
5-gram & 60.5 & 47.9 & 71.9 & 64.4 & 68.5 & 70.0 & 36.9 & 58.1 & 79.5 & 53.7 & 45.5 & 53.5 & 60.3\\
LSTM & 68.9 & 91.7 & 73.2 & 73.5 & 67.0 & 85.4 & 67.6 & 72.5 & 89.1 & 42.9 & 51.7 & 64.5 & 80.1\\
TXL & 68.7 & 94.1 & 69.5 & 74.7 & 71.5 & 83.0 & 77.2 & 64.9 & 78.2 & 45.8 & 55.2 & 69.3 & 76.0\\
GPT-2 & 80.1 & 99.6 & 78.3 & 80.1 & 80.5 & 93.3 & 86.6 & 79.0 & 84.1 & 63.1 & 78.9 & \bf 71.3 & 89.0\\
BERT\subtxt{BASE} & 84.2 & 97.0 & 80.0 & \bf 82.3 & 79.6 & \bf 97.6 & 89.4 & 83.1 & \bf 96.5 & 73.6 & \bf 84.7 & 71.2 & \bf 92.4\\
RoBERTa\subtxt{BASE} & \bf 85.4 & 97.3 & \bf 83.5 & 77.8 & \bf 81.9 & 97.0 & 91.4 & \bf 90.1 & 96.2 & \bf 80.7 & 81.0 & 69.8 & 91.9\\
Human & 88.6 & 97.5 & 90.0 & 87.3 & 83.9 & 92.2 & 85.0 & 86.9 & 97.0 & 84.9 & 88.1 & 86.6 & 90.9\\
1B-1 & 82.3 & 97.7 & 80.7 & 77.3 & 80.7 & 95.8 & 91.6 & 83.1 & 92.5 & 69.7 & 79.9 & 68.7 & 89.4\\
1B-2 & 81.0 & 97.5 & 79.1 & 78.3 & 79.4 & 96.0 & \bf 92.2 & 82.1 & 94.8 & 63.4 & 81.2 & 61.7 & 89.6\\
1B-3 & 82.0 & \bf 98.6 & 79.3 & 78.5 & 77.2 & 95.3 & 91.2 & 83.1 & 94.8 & 66.5 & 82.6 & 70.5 & 89.5\\
100M-1 & 76.3 & 93.9 & 74.6 & 72.7 & 77.0 & 93.2 & 89.9 & 74.3 & 89.9 & 60.6 & 76.6 & 61.6 & 78.1\\
100M-2 & 79.7 & 97.2 & 79.1 & 75.4 & 79.6 & 94.5 & 91.6 & 78.8 & 92.7 & 63.0 & 77.2 & 64.7 & 87.5\\
100M-3 & 79.1 & 95.8 & 76.9 & 76.0 & 75.4 & 95.6 & 93.7 & 76.8 & 93.9 & 62.5 & 80.2 & 60.9 & 86.9\\
10M-1 & 72.0 & 88.0 & 70.3 & 74.0 & 70.3 & 90.0 & 83.7 & 66.8 & 89.6 & 51.5 & 71.3 & 62.9 & 74.5\\
10M-2 & 72.6 & 91.1 & 70.1 & 71.6 & 70.7 & 91.6 & 86.0 & 67.3 & 84.3 & 53.6 & 75.6 & 58.6 & 77.0\\
10M-3 & 71.4 & 91.4 & 71.1 & 71.4 & 66.4 & 90.5 & 85.3 & 65.8 & 91.3 & 46.8 & 69.1 & 62.3 & 81.1\\
1M-1 & 58.5 & 67.9 & 60.4 & 58.5 & 59.4 & 59.5 & 54.6 & 61.6 & 78.1 & 50.8 & 54.2 & 64.8 & 52.5\\
1M-2 & 58.5 & 66.0 & 60.0 & 57.8 & 58.8 & 61.1 & 55.7 & 61.5 & 78.6 & 48.7 & 55.0 & 65.5 & 54.2\\
1M-3 & 58.7 & 68.4 & 60.3 & 57.5 & 59.1 & 61.3 & 55.1 & 61.2 & 77.7 & 48.5 & 56.6 & 67.2 & 52.9\\ \hline 
\end{tabular}}
\caption{BLiMP results. 5-gram, LSTM, TXL, GPT-2 scores come from \citet{warstadt2020blimp}. BERT\subtxt{BASE} scores come from \citet{salazar-etal-2020-masked}.}
\label{tab:mlms-blimp}
\end{table*}

\begin{figure*}
    \centering
    \includegraphics[width=\textwidth]{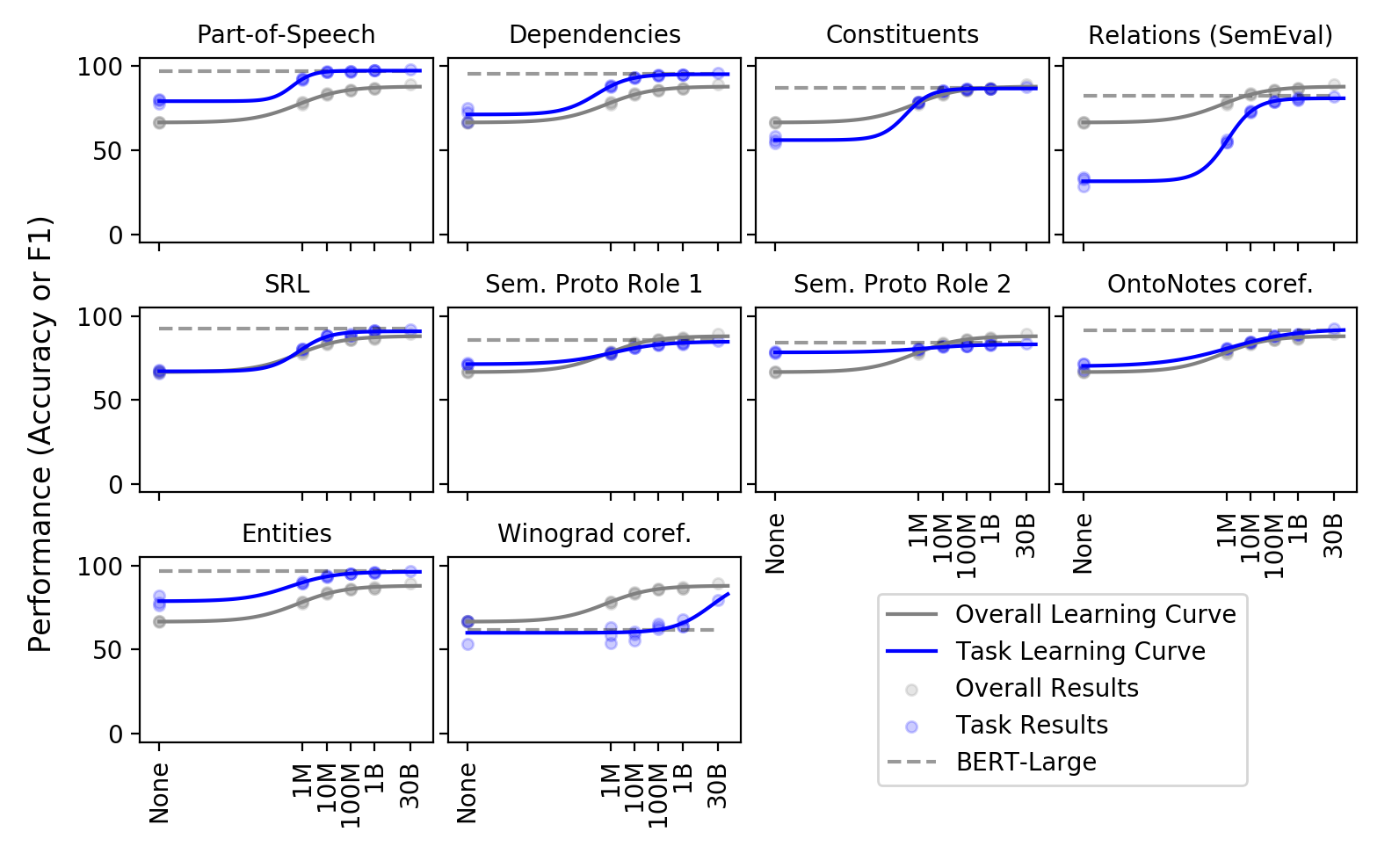}
    \caption{Our absolute edge probing dev set results (not normalized) compared to BERT\subtxt{LARGE} test set results from \citet{tenney2019you}.}
    \label{fig:my_label}
\end{figure*}


\begin{figure*}
    \centering
    \includegraphics[width=\textwidth]{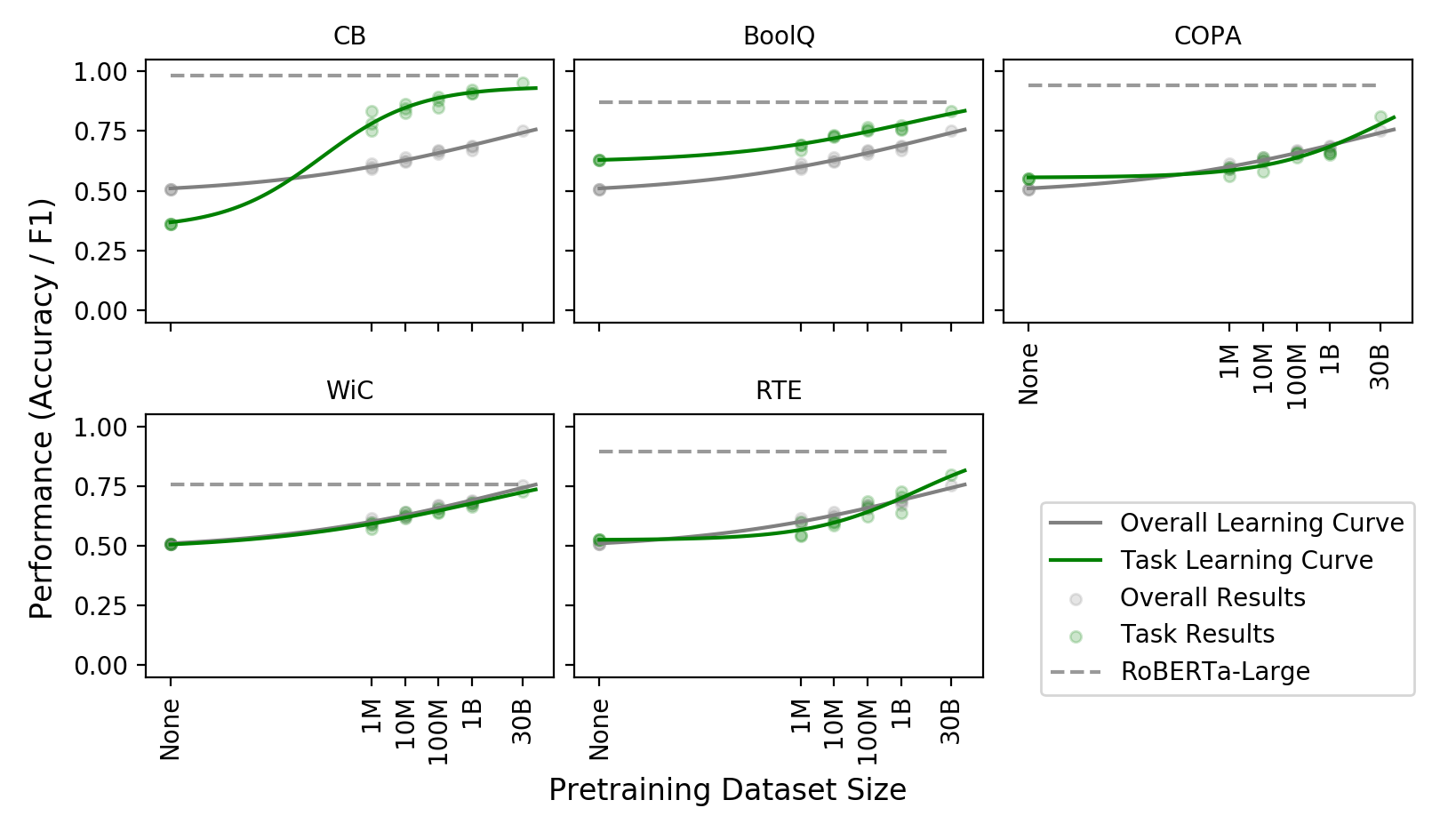}
    \caption{Our absolute SuperGLUE results (not normalized) compared to RoBERTa\subtxt{LARGE} results from \citet{liu2019roberta}. 
    }
    \label{fig:my_label}
\end{figure*}

\end{document}